\documentclass{article}

\usepackage{graphicx} 

\usepackage[preprint]{neurips_2024}

\usepackage[utf8]{inputenc} 
\usepackage[T1]{fontenc}    
\usepackage{url}            
\usepackage{booktabs}       
\usepackage{amsfonts}       
\usepackage{nicefrac}       
\usepackage{microtype}      
\usepackage{multirow} 
\usepackage{fontawesome}
\usepackage{amsmath}

\usepackage[table]{xcolor}
\usepackage{wrapfig}       
\usepackage{lipsum}

\title{ ManipLVM-R1: Reinforcement Learning for Reasoning in Embodied Manipulation with Large Vision-Language Models}

\newcommand{\model}{ManipLVM-R1}
\definecolor{DarkGreen}{RGB}{0,100,0}
\definecolor{DarkPurple}{RGB}{128,0,128}

\author{
Zirui Song$^{1}$\thanks{Equal Contribution} \ \   
Guangxian Ouyang$^{1*}$  \ \   
Mingzhe Li$^{2}$ \ \  
Yuheng Ji$^{3}$ \ \ 
Chenxi Wang$^{1}$ \ \  
\textbf{Zixiang Xu}$^{1}$ \\ 
\textbf{Zeyu Zhang}$^{4}$ \ \ \ 
\textbf{Xiaoqing Zhang}$^{5}$ \ \ \
\textbf{Qian Jiang}$^{1}$ \ \ \    
\textbf{Zhenhao Chen}$^{1}$ \ \ \
\textbf{Zhongzhi Li}$^{3}$ \\
\textbf{Rui Yan}$^{6}$ \ \ \
\textbf{Xiuying Chen}$^{1}$\thanks{Corresponding Author.} 
\\
\textsuperscript{1} Mohamed bin Zayed University of Artificial Intelligence \\
\textsuperscript{2} ByteDance 
\textsuperscript{3} Institute of Automation, Chinese Academy of Sciences \\
\textsuperscript{4} The Australia National University
\textsuperscript{5} Renmin University of China \\
\textsuperscript{6} Wuhan University
}

\begin{document}

\maketitle

\begin{abstract}

Large Vision-Language Models (LVLMs) have recently advanced robotic manipulation by leveraging vision for scene perception and language for instruction following.
However, existing methods rely heavily on costly human-annotated training datasets, which limits their generalization and causes them to struggle in out-of-domain (OOD) scenarios, reducing real-world adaptability.
To address these challenges, we propose \model, a novel reinforcement learning framework that replaces traditional supervision with Reinforcement Learning using Verifiable Rewards (RLVR).
By directly optimizing for task-aligned outcomes, our method enhances generalization and physical reasoning while removing the dependence on costly annotations.
Specifically, we design two rule-based reward functions targeting key robotic manipulation subtasks: an Affordance Perception Reward to enhance localization of interaction regions, and a Trajectory Match Reward to ensure the physical plausibility of action paths. These rewards provide immediate feedback and impose spatial-logical constraints, encouraging the model to go beyond shallow pattern matching and instead learn deeper, more systematic reasoning about physical interactions.
Experimental results show that \model{} achieves substantial performance gains across multiple manipulation tasks, using only 50\% of the training data while achieving strong generalization to OOD scenarios.
We further analyze the benefits of our reward design and its impact on task success and efficiency.
\end{abstract}

\section{Introduction}
Large Vision Language Models (LVLMs) are accelerating progress toward artificial general intelligence (AGI) by combining self-supervised learning with large-scale multimodal data~\citep{song2025geolocation,xie2025medtrinity25mlargescalemultimodaldataset}. They exhibit strong visual perception and language understanding, and have achieved success in tasks such as visual question answering~\citep{vqa,cai2024benchlmm,internvl}, image captioning~\citep{mplug,hao2023mixgen,blip2}, and mobile assistants~\citep{song2024mmac,li2024appagent}, demonstrating robust cross-modal reasoning.

However, their use in robotics—especially in manipulation tasks involving complex physical interaction—remains limited due to a lack of fine-grained control and physical reasoning. Most existing work focuses on high-level planning~\citep{Rt-h,inner,song2024hazards}, action sequencing~\citep{Saycan,Rt-2}, or visual reasoning~\citep{reflect,selfcorrect,code_as_monitor,azzolini2025cosmos,zhang2025embodied,zhao2025embodied}, with limited attention to low-level decision-making and physical dynamics, making it difficult to tackle real-world robotic challenges.
\begin{figure}[h]
    \centering
    \includegraphics[width=1\linewidth]{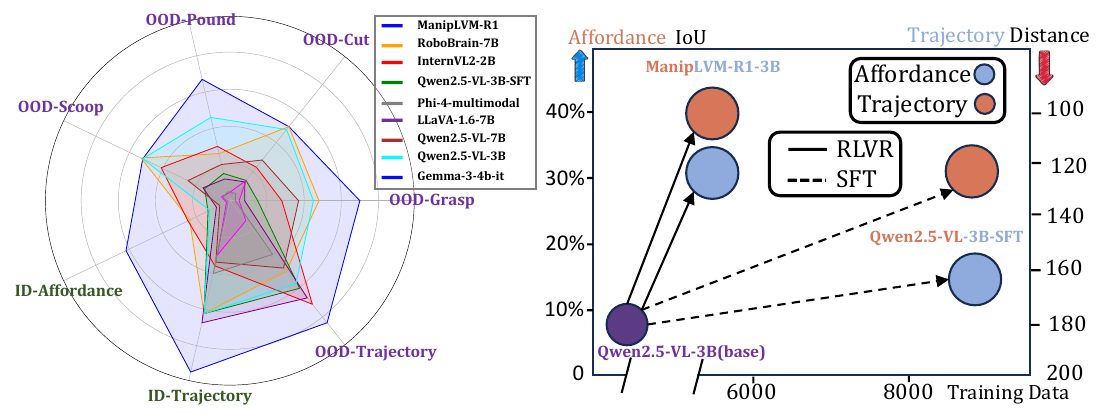}
    \caption{
    \textbf{Left:} Comparative evaluation of \model{} against baselines on \textcolor{DarkGreen}{In Domain (ID)} dataset and \textcolor{DarkPurple}{Out of Domain (OOD)} dataset.
    \textbf{Right:} Leveraging our proposed RLVR method, \model{} outperforms supervised fine-tuning on both affordance perception and trajectory prediction, using only 50\% of the training data.
}
\label{fig:intro}
\vspace{-6mm}
\end{figure}

Among existing studies, RoboBrain~\citep{ji2025robobrain} is most closely related to our work, as it also leverages LVLMs for robotic manipulation through reasoning-augmented training.
It constructs a supervised fine-tuning (SFT) dataset by combining expert demonstrations with language model generated reasoning.
While this approach enhances alignment with task instructions, SFT-based methods suffer from two major limitations: the high cost of collecting high quality annotations, and limited generalization due to their reliance on fixed, supervised data. 
When transferred to new datasets or applied in out-of-distribution scenarios, their performance degrades significantly.
These limitations reduce the scalability and adaptability of such systems, particularly when deployed in diverse or previously unseen environments.

Given its success in domains such as mathematics and programming~\citep{lambert2024t,team2025kimi,liu2024vuldetectbenchevaluatingdeepcapability,huang2025trustworthiness,wang2025trusteval,huang2025breaking,chen2025unveiling,han2024medinst,wang2025word}, Reinforcement Learning with Verifiable Rewards(RLVR) offers a promising direction for addressing the challenges of data dependency and limited generalization in robotic manipulation. 
However, applying RLVR to robotic manipulation also presents two key challenges. First, reward signals in manipulation tasks are often sparse and delayed, making it difficult to deliver timely and informative feedback for effective optimization. Second, existing reward designs typically lack spatial and logical constraints, causing models to overfit to superficial patterns rather than learning grounded physical reasoning and systematic understanding.

To address these challenges, we propose \model, a novel reinforcement learning framework that leverages RLVR to provide timely and verifiable feedback for robotic manipulation. 
Instead of relying on sparse and delayed signals, \model{} introduces structured, task-aligned rewards for two key subtasks: affordance perception and trajectory prediction.
For \textbf{\textit{affordance perception}}, \model{} combines response format validation with IoU-based spatial localization to produce clear and prompt reward signals. 
For \textbf{\textit{trajectory prediction}}, it incorporates a mixture of trajectory distance metrics, endpoint distance, and trajectory length constraints to evaluate both the quality and feasibility of action sequences.
To overcome the limitations of shallow reward designs, \model{} embeds spatial structure and path consistency into its objectives, encouraging the model to develop a systematic understanding of physical interactions rather than relying on superficial patterns.
Experimental results demonstrate that \model{} substantially improves performance across diverse robotic manipulation tasks, including a 144\% increase in Intersection over Union (IoU) over the strongest baseline in affordance perception and a 12.5\% improvement in trajectory prediction—achieved using only 50\% of the training data.
Furthermore, in our OOD test, \model{} surpassed all evaluated open-source and supervised fine-tuned models, including Qwen2.5-VL-32B-Instruct and Robobrain-7B~\citep{ji2025robobrain}. 
Notably, we provide a detailed analysis of our reward design and demonstrate its impact on task success and efficiency.

Our contributions are threefold. 
First, we propose a novel adaptation of the RLVR paradigm for robotic manipulation, introducing a verifiable reward-based training framework that requires no human annotations.
Second, we design structured reward mechanisms tailored to affordance perception and trajectory prediction, guiding the model to generate executable responses based on spatial precision and path alignment. Third, we conduct comprehensive evaluations on multiple challenging manipulation tasks, demonstrating that \model{} outperforms existing supervised fine-tuning approaches in terms of sample efficiency, task success rate, and generalization capability.

\section{Related Work}

\subsection{ RLVR for Large Vision Language Models}
LVMs have demonstrated remarkable reasoning capabilities across various visual tasks~\citep{LLaVa,llava-1.5,li2024llava,qwen2vl,Qwen,wang2024qwen2}.
Recently, several works have explored the use of RLVR to further enhance visual reasoning performance.
Vision-R1~\citep{huang2025vision} combines a human-annotation-free cold-start math reasoning dataset with a Progressive Thinking Suppression Training (PTST) strategy, achieving performance on par with much larger models.
LMM-R1~\citep{peng2025lmm} proposes a two-stage RLVR framework that first improves foundational reasoning using text data, then generalizes to vision-based and agent-centric reasoning domains.
VLM-R1~\citep{shen2025vlm} adapts R1-style reinforcement learning to visual grounding tasks, demonstrating improved performance and generalization in open-vocabulary object detection.
Despite these promising results, RLVR remains underexplored in the context of robotic manipulation. To bridge this gap, we propose \model—a novel RLVR-based framework designed to enhance both reasoning and manipulation capabilities.

\subsection{Robot Manipulation}
Traditional state-based reinforcement learning methods~\citep{geng2023rlafford,andrychowicz2020learning} laid the foundation for early research in robotic manipulation. However, these methods often rely on low-dimensional sensor states and struggle to generalize to high-dimensional, real-world visual inputs. In recent years, research has increasingly shifted toward vision-centric manipulation tasks, leveraging the strong reasoning capabilities of Large Language Models (LLMs) to enhance the intelligence and generalization of manipulation strategies~\citep{Rt-1,Rt-2,wan2023unidexgrasp++,wang2023sparsedff,li2024manipllm,Robomamba}.
Specifically, VoxPoser~\citep{VoxPoser} utilizes LLMs to generate 3D value maps through Vision-Language Models (VLMs), enabling zero-shot synthesis of manipulation trajectories from language instructions. RoboFlamingo~\citep{li2023vision} fine-tunes models on vision-language manipulation datasets to perform language-conditioned manipulation tasks. ManipLLM~\citep{li2024manipllm} further incorporates the Chain-of-Thought (CoT) paradigm, fine-tuning adapters to integrate object understanding, affordance reasoning, and pose prediction into an interpretable, object-centric manipulation framework.
Building on these approaches, OpenVLA~\citep{OpenVLA} and RoboMamba~\citep{Robomamba} construct more fine-grained CoT datasets and employ supervised fine-tuning strategies to further improve manipulation performance. In parallel, Embodied Reasoner~\citep{zhang2025embodied}, Cosmos-Reason1~\citep{azzolini2025cosmos}, and RoboBrain~\citep{ji2025robobrain} focus on enhancing long-horizon reasoning and improving interpretability and logical consistency in the decision-making process.
Despite recent advances, most manipulation VLMs still depend on large-scale, high-quality CoT datasets. 
In contrast, our proposed \model{} relies on a small set of labeled examples and adapts the RLVR paradigm to stimulate the inherent reasoning capabilities of the base model, enabling low-supervision and high-generalization manipulation.




\section{Method}

\begin{figure*}[t]
\centering
\includegraphics[scale=0.26]{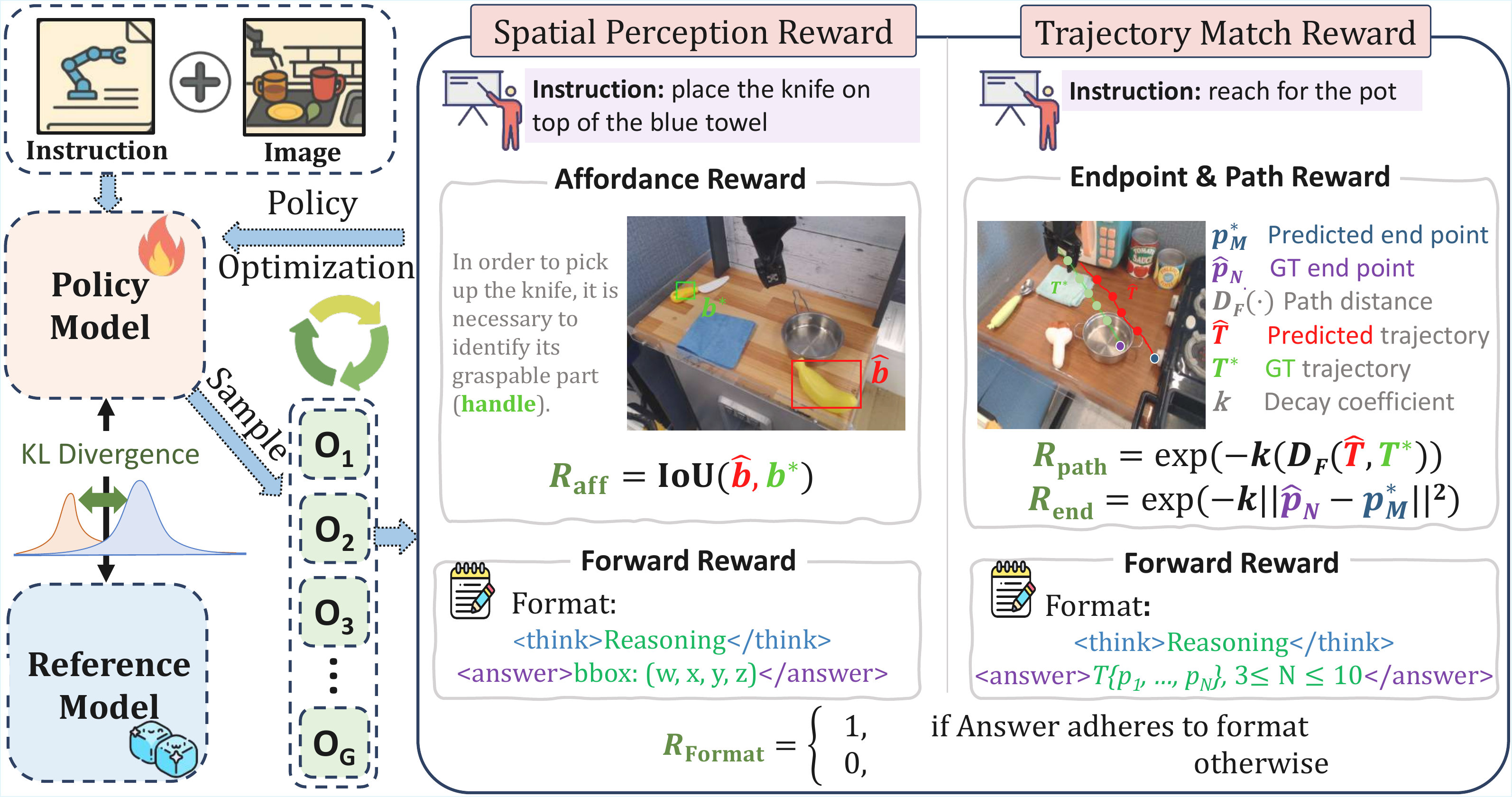}
\caption{
Overview of the proposed \model{} framework. Given an image and an instruction, the policy model generates multiple responses, which are optimized using structured rewards: the Affordance Perception Reward for affordance detection, and the Trajectory Match Reward based on multiple distance metrics, path similarity, and endpoint accuracy.
}
\label{fig:model}
\end{figure*}

In this section, we first present the preliminaries of RLVR. Subsequently, we introduce the training framework for \model{} and elucidate the rationale behind its design. The visualization of \model{} is shown in Figure~\ref{fig:model}.

\subsection{Preliminary of Reinforcement Learning with Verifiable Rewards}

RLVR \citep{lambert2024t,DeepSeek-R1,team2025kimi} is a training paradigm designed to enhance language models in tasks where correctness can be objectively verified, such as mathematics and coding.
Unlike reinforcement learning from human feedback~\citep{liu2024skywork,zang2025internlm}, RLVR derives its reward signal directly from a verifiable reward function.
Given the input instruction $q$, the policy model $\pi_\theta$ generates responses $o$ and receives the verifiable reward.
Concretely, RLVR optimizes the following objective:
\begin{align}
\max_{\pi_\theta} \mathbb{E}_{o \sim \pi_\theta(q)} \left[ R_{\text{RLVR}}(q, o) \right]  = \left[ R(q, o) - \beta \, \mathrm{KL} \left[ \pi_\theta(o|q) \,\|\, \pi_{\text{ref}}(o|q) \right] \right],
\label{kl}
\end{align}
where $\pi_{\text{ref}}$ is the reference model before optimization, $R$ is the verifiable reward function, and $\beta$ is the hyperparameter to control the KL-divergence~\citep{hershey2007approximating}.
The verifiable reward function $R$ takes the instruction and output pair $(q, o)$ as inputs, and checks if the ground-truth answer remains the same as the prediction $o$:
\begin{equation}
R(q, o) =
\begin{cases}
1, & \text{if } o = \text{ground truth}, \\
0, & \text{otherwise}.
\end{cases}
\end{equation}

\subsection{\model}
While RLVR has shown promise in domains like mathematics and programming, its application to robotic manipulation remains limited.
This is primarily due to two challenges: (1) sparse and delayed reward signals in manipulation tasks, which hinder effective learning and optimization, and (2) the lack of spatial-logical constraints in reward design, causing models to rely on superficial pattern matching rather than learning grounded physical reasoning.
To address these issues, we propose \model, an RLVR-based framework that incorporates structured reward design and stable policy optimization.
\model{} decomposes the overall manipulation task into two distinct subtasks: affordance perception and trajectory prediction.
It trains separate models for each subtask, enabling more focused learning and better generalization in physically grounded environments.

\paragraph{Policy Samples} For a given input state $(x,q)$, where $x$ is the visual encoding of the input image and $q$ is the textual encoding of the instruction, \model{} first generates $G$ distinct responses $\{o_1,o_2,...,o_G\}$ from the current policy $\pi_{\theta}$. 
Different from other work, to adapt this methodology for robotic manipulation tasks involving affordance perception and trajectory prediction, \model{} incorporates distinct sets of reward functions: \textit{Affordance Perception Reward} and \textit{Trajectory Match Reward}.
For brevity, we omit the response index in the following introduction.

\paragraph{\textit{Affordance Perception Reward}}  
For the affordance perception task, the Affordance Perception Reward $R_{\text{spatial}}$ combines two components: a format reward $R_{\text{format}}$ and an affordance reward $R_{\text{aff}}$.  
The format reward \( R_{\text{format}} \) ensures that the model outputs follow a structured format—reasoning enclosed in `<think></think>' and the final answer enclosed in `<answer></answer>', with the `<answer>' field containing the coordinates of a predicted bounding box. 
This structure not only facilitates reliable reward extraction but also guides the model to organize its output in a way that aligns with downstream evaluation and training.
Meanwhile, the affordance reward \( R_{\text{aff}} \) focuses on the core objective of affordance perception: accurately identifying where an object interaction can occur. 
It evaluates spatial accuracy by comparing the predicted bounding box \( b^* \) with the ground-truth box \( \hat{b} \) using the Intersection-over-Union (IoU) metric.  
By optimizing \( R_{\text{aff}} = \text{IoU}(b^*, \hat{b}) \), the model is encouraged to generate spatially precise and semantically grounded predictions that directly support successful interaction planning.
Taken together, the Affordance Perception Reward is defined as:
\begin{equation}
 R_{\text{spatial}} = R_{\text{format}} + R_{\text{aff}}.   
\end{equation}

\paragraph{\textit{Trajectory Match Reward}}
For the trajectory prediction task, the model is expected to output a sequence of 2D coordinates, denoted as \( \hat{T} = \{\hat{p}_1, \hat{p}_2, ..., \hat{p}_N\} \), where each \( \hat{p}_i \in \mathbb{R}^2 \) represents an \((x, y)\) position in image space, with the origin at the top-left corner.
The predicted trajectory is evaluated against a ground-truth sequence \( T^* = \{p^*_1, p^*_2, ..., p^*_M\} \) using three components: a format reward \( R_{\text{format}} \), a path similarity reward \( R_{\text{path}} \), and an endpoint distance reward \( R_{\text{end}} \).

The format reward \( R_{\text{format}} \) verifies structural correctness, including proper use of `<think>` and `<answer>` tags. 
Unlike in affordance perception, it enforces a trajectory-specific constraint: the predicted numbers must fall within a valid range (e.g., 3–10), ensuring both conciseness and interpretability.

To encourage the model to generate physically plausible and goal-aligned trajectories, we design the path similarity reward $R_{\text{path}}$.
The motivation behind this design is that many supervised methods, such as behavioral cloning~\citep{nazeri2021exploring}, focus primarily on endpoint accuracy while overlooking the quality and feasibility of the trajectory itself.
In robotic manipulation tasks, the shape, continuity, and alignment of the trajectory with respect to the ground truth are critical for ensuring robust and safe control strategies.
To capture these aspects, we incorporate multiple geometric similarity metrics that evaluate how well the predicted trajectory matches the ground truth from different perspectives.
$R_{\text{path}}$ is a weighted aggregation of rewards derived from three distinct metrics: Discrete Fréchet Distance (DFD)~\citep{eiter1994computing}, Hausdorff Distance (HD)~\citep{huttenlocher1993comparing}, and Root Mean Square Error (RMSE).
The reward for each metric ($R_{\text{DFD}}, R_{\text{HD}}, R_{\text{RMSE}}$) is obtained by normalizing the respective distance value to a score between 0 and 1.
$R_{\text{DFD}}$ quantifies overall shape similarity and temporal alignment between trajectories;
$R_{\text{HD}}$ captures the maximum pointwise deviation;
$R_{\text{RMSE}}$ measures the average pointwise discrepancy.
The final $R_{\text{path}}$ is as:
\begin{equation}
    R_{\text{path}}=R_{\text{DFD}} +  R_{\text{HD}} +R_{\text{RMSE}}.
\end{equation}


The endpoint distance reward $R_{\text{end}}$ measures the proximity between the final predicted point $\hat{p}_N$ and the ground-truth endpoint $p_M$ using Euclidean distance: $R_{\text{end}}=\exp(-k||\hat{p_N}-p^*_M||^2)$.
The overall trajectory match reward is defined as the sum of the three components: 
\begin{equation} R_{\text{trajectory}}= R_{\text{format}}+ R_{\text{path}} + R_{\text{end}}.
\label{rrr}
\end{equation}

\paragraph{Policy Update}  Inspired by Group Relative Policy Optimization (GRPO)~\citep{DeepSeek-R1}, we select multiple responses from the current policy as candidate actions. 
Depending on the task type, either \(R_{\text{spatial}}\) or \(R_{\text{trajectory}}\) assigns a scalar reward \(r_i\) to each response, resulting in a reward set \(\{r_1, r_2, \ldots, r_G\}\).  
To assess the quality of each response relative to others, we normalize the rewards by computing the mean and standard deviation:
\begin{equation}
A_i = \frac{r_i - \mathrm{mean}(\{r_1, \ldots, r_G\})}{\mathrm{std}(\{r_1, \ldots, r_G\})},
\end{equation}
where \(A_i\) denotes the advantage of the \(i\)-th response.  
These advantages are then used to update the policy, increasing the likelihood of high-quality responses while suppressing lower-quality ones. To ensure stability during training, the update is constrained by minimizing the KL divergence between the updated policy and a reference model: $\mathrm{KL} \left[ \pi_\theta(o|q) \,\|\, \pi_{\text{ref}}(o|q) \right]$.

\begin{wraptable}{r}{0.7\textwidth}
\vspace{-2mm}
\caption{Comparison of model performance on In-Domain dataset. \textbf{Bolded} values indicate the best performance, and \underline{underlined} values indicate the second best. Task-specific metrics are color-coded: \textcolor{red!40}{red} for Affordance Perception and \textcolor{green!80}{green} for Trajectory Prediction.}
\label{table1}
\resizebox{0.7\textwidth}{!}{
\begin{tabular}{r|c|cccc}
\hline
\multirow{2}{*}{Method} & 
\multicolumn{5}{c}{\cellcolor{cyan!20}{In-Domain}} \\
& \cellcolor{red!20}{IoU $\uparrow$} & \cellcolor{green!20}{DFD $\downarrow$} & \cellcolor{green!20}{HD $\downarrow$} & \cellcolor{green!20}{RMSE $\downarrow$} & \cellcolor{green!20}{Avg$\downarrow$} \\
\hline

\multicolumn{1}{l|}{\cellcolor[HTML]{F5F5F5}\textit{\textbf{Open-source Models}}} & \cellcolor[HTML]{F5F5F5} & \multicolumn{4}{c}{\cellcolor[HTML]{F5F5F5}}  \\
Phi-4-multimodal-Instruct & 0.58 & 243.92 & 224.73 & 189.27 & 228.21\\
Gemma-3-4b-it & 0.91 & 266.62 & 243.19 & 210.48 &  240.10 \\
Gemma-3-27b-it & 1.32 & 257.42 & 230.29 & 184.47& 224.00\\
Gemma-3-12b-it & 1.18 & 206.72 &190.64  &154.96 & 184.10\\
Qwen2.5-VL-3B-Instruct & 6.15 & 202.86 & 179.12 & 144.14 & 175.37 \\
Qwen2.5-VL-7B-Instruct & 2.98 & 262.80 & 243.03 & 190.81 & 232.21 \\
Qwen2.5-VL-32B-Instruct & 7.40 & \textbf{125.54} & \textbf{113.00} & \textbf{85.05} & \textbf{107.86}\\
\hline
\multicolumn{1}{l|}{\cellcolor[HTML]{FFE4B5}\textit{\textbf{Supervised Fine-Tuning}}} & \cellcolor[HTML]{FFE4B5} & \multicolumn{4}{c}{\cellcolor[HTML]{FFE4B5}} \\
LLaVA-1.6-7B & 3.98 & 184.40 & 178.00 & 133.28 & 165.23\\
InternVL2-2B & 6.74 & 250.20 & 239.34 & 194.74 & 228.09\\
Qwen2.5-VL-3B-Instruct & \underline{12.69} & 147.38 & 138.90 & 94.03 & 126.77 \\
RoboBrain-7B & 11.79 & 156.10 & 136.52 & 106.71 & 133.11 \\
\hline

\multicolumn{1}{l|}{\cellcolor[HTML]{FFE4B5}\textit{\textbf{Our Proposed Model}}} & \cellcolor[HTML]{FFE4B5} & \multicolumn{4}{c}{\cellcolor[HTML]{FFE4B5}} \\
\model-3B & \textbf{31.0} & \underline{134.18} & \underline{111.14} & \underline{87.28} & \underline{110.87}\\
\hline
\end{tabular}
}
\vspace{-\intextsep} 
\end{wraptable}

\section{Experiment}

\subsection{Baselines}




To establish robust performance benchmarks, we curated a comprehensive suite of baseline models. 
Specifically, we selected three of the most recently released, highest-performing open-source multimodal language model series: Gemma-3-4B-it, Gemma-3-12B-it, Gemma-3-27b-it~\citep{team2025gemma}, Phi-4-multimodal-Insturct~\citep{abouelenin2025phi}, and Qwen2.5-VL-3B-Instruct, Qwen2.5-VL-7B-Instruct, Qwen2.5-VL-32B-Instruct~\citep{qwen2_5}. 
We implemented few-shot prompting to ensure they possessed fundamental perceptual capabilities.

Furthermore, to validate the efficacy of our proposed training framework, we selected three models as baselines for supervised fine-tuning. These include InternVL2-2B~\citep{internvl}, Qwen2.5-VL-3B-Insturct~\citep{qwen2_5}, LLaVA-1.6-7B~\citep{LLaVa} and RoboBrain-7B~\citep{ji2025robobrain}. 
Notably, Qwen2.5-VL-3B-Instruct serves as our base model. RoboBrain is specifically designed for affordance perception and trajectory prediction, which was fine-tuned from its base model, LLaVA-1.6-7B, using an extensive, meticulously prepared synthetic chain-of-thought dataset. All baseline models were fully fine-tuned on 100\% of the ShareRoBot training set, while our proposed model, \model, used only 50\% of the data.


\subsection{Dataset Setting}
\paragraph{In Domain Dataset} 
Both our baseline models and the proposed \model{} were trained on the ShareRobot~\citep{ji2025robobrain} dataset, a large-scale dataset designed to enhance affordance perception and trajectory prediction.
The data was curated from Open X-Embodiment~\citep{o2024open} using strict quality criteria and human verification. 
The dataset includes fine-grained, multi-dimensional annotations such as low-level planning instructions, object affordances, and end-effector trajectories.
ShareRobot comprises over 1M planning QA pairs, 6.5k affordance-annotated images, and 6.8k trajectory-annotated images, covering diverse scenarios (102 scenes, 12 embodiments). 
Notably, our proposed model, \model, was trained using only 50\% of the training set. 
In contrast, all SFT baseline models used for comparison were trained on the entire training dataset.

\paragraph{Out of Domain Dataset}
To evaluate the generalization ability of our model, we also conducted OOD experiments.
For the affordance perception task, we employed subsets from the UMD Part Affordance dataset \citep{UMD} as an out-of-domain benchmark for our affordance task evaluation.  
The UMD dataset consists of RGB-D images and 3D point clouds representing 105 tools commonly used in gardening, kitchen, and workshop environments. Specifically, we selected the `grasp', `cut', `pound', and `scoop' categories. From these selected categories, we randomly sampled a total of 1200 data pairs/samples, ensuring an equal distribution across the four categories, to constitute our OOD test benchmark.

For the trajectory prediction task, we designated a randomly selected subset of 500 samples from the validation data of LLARVA's pre-training dataset, VAIT~\citep{niu2024llarva}, as our OOD test set. VAIT, derived from the diverse Open X-Embodiment dataset, encompasses image-visual trace pairs across numerous robotic scenarios. Given that VAIT is a large-scale pre-training dataset, our manual inspection revealed the presence of some noise and inaccuracies in annotations. To ensure rigorous testing, we visualized all trajectories within this randomly selected subset and conducted human verification of the data annotations. For instances exhibiting conspicuous deviations, we performed manual corrections.

\begin{table}
\caption{Comparison of model performance in out-of-domain (OOD) dataset. 
\textbf{Bolded} values signify optimal outcomes and \underline{underlined} values indicate suboptimal outcomes. \textcolor{red!40}{Red} cell for the Affordance Perception task and \textcolor{green!80}{green} cell for the trajectory prediction task.}
\label{table2}
\resizebox{\textwidth}{!}{
\begin{tabular}{r|cccc|cccc}
\hline
\multirow{3}{*}{Method} & 
\multicolumn{8}{c}{\cellcolor{cyan!20}{Out of Domain}} \\
& \multicolumn{4}{c|}{\cellcolor{red!20}{\textbf{UMD Affordance}}} 
& \multicolumn{4}{c}{{\cellcolor{green!20}\textbf{VAIT Trajectory}}} \\ 

& \cellcolor{red!20}{Grasp-IoU $\uparrow$} & \cellcolor{red!20}{Cut-IoU $\uparrow$} & \cellcolor{red!20}{Pound-IoU $\uparrow$} & \cellcolor{red!20}{Scoop-IoU $\uparrow$} & \cellcolor{green!20}{DFD$\downarrow$} & \cellcolor{green!20}{HD$\downarrow$} & \cellcolor{green!20}{RMSE $\downarrow$} & \cellcolor{green!20}{Avg$\downarrow$} \\
\hline

\multicolumn{1}{l|}{\cellcolor[HTML]{F5F5F5}\textit{\textbf{Open-source Models}}} & \cellcolor[HTML]{F5F5F5} & \multicolumn{7}{c}{\cellcolor[HTML]{F5F5F5}} \\
Phi-4-multimodal-Instruct &1.49 &2.53 &2.44 & 2.22 & 240.18 & 235.44& 202.69 & 226.10\\
Gemma-3-4b-it & 2.29 & 7.04& 2.24& 2.26 & 295.48 &289.41 & 232.21 & 272.31\\
Gemma-3-12b-it & 4.14 & 5.25& 4.36& 4.66 & 204.94& 200.89 & 175.42 & 193.75 \\
Gemma-3-27b-it & 10.29 & 2.08&6.27 & 4.05 & 273.86 &268.63 & 209.52 & 250.67 \\
Qwen2.5-VL-3B-Instruct &22.48 & 24.50&22.77 &26.09 &211.80 &205.04 &140.69 & 250.67 \\
Qwen2.5-VL-7B-Instruct & 18.64 & 14.19 &10.07 & 12.31 & 228.04&222.92 &170.94 & 207.30\\
Qwen2.5-VL-32B-Instruct &\underline{24.67} & \underline{25.84} & \textbf{24.04} & 25.99 & 182.73 & 176.51 & \underline{133.17} & 164.14\\
\hline
\multicolumn{1}{l|}{\cellcolor[HTML]{FFE4B5}\textit{\textbf{Surpervised Fine-Tuning}}} & \cellcolor[HTML]{FFE4B5} & \multicolumn{4}{c}{\cellcolor[HTML]{FFE4B5}} & \multicolumn{3}{c}{\cellcolor[HTML]{FFE4B5}}  \\
LLaVA-1.6-7B & 3.89 & 6.34 &  6.05& 7.31  & 170.88 &167.10 & 160.79 & 166.25\\
Qwen2.5-VL-3B-Instruct & 7.39 & 7.11 & 7.50& 7.81& 186.48&184.95 &166.92 & 179.45 \\
InternVL2-2B& 13.93&11.52 &  15.06 &20.48 & \underline{165.98} & \underline{160.87} & 145.64 & \underline{157.50}\\
RoboBrain-7B & 23.86 & 25.37 & 12.75 & \underline{26.35} & 220.94 & 214.14 & 173.02 & 202.70\\
\hline
\multicolumn{1}{l|}{\cellcolor[HTML]{FFE4B5}\textit{\textbf{Our Proposed Model}}} & \cellcolor[HTML]{FFE4B5} & \multicolumn{7}{c}{\cellcolor[HTML]{FFE4B5}} \\
    \model-3B & \textbf{34.65} & \textbf{25.58} & \underline{23.50} & \textbf{28.27} & \textbf{146.82} & \textbf{140.52} & \textbf{108.64} & \textbf{131.99} \\
\bottomrule
\end{tabular}
}
\vspace{-3mm} 
\end{table}


\subsection{Metric Setting}
\textbf{For Affordance Perception:} We use Intersection over Union (IoU) as the metric to quantify the spatial accuracy of the predicted affordance region by measuring the overlap between the predicted region and the ground truth annotation. A higher IoU value indicates better spatial alignment between the prediction and the ground truth. 
\textbf{For Trajectory Prediction:} We evaluate the concordance between ground truth and predicted trajectories. Following previous work~\citep{chao2021dexycb,qwen2vl,ji2025robobrain}, the trajectory is represented as ordered sets of two-dimensional waypoints, normalized to a range of [0, 1000). The assessment of trajectory similarity incorporates three distinct metrics: the Discrete Fréchet Distance (DFD), the Hausdorff Distance (HD), and the Root Mean Square Error (RMSE). The DFD metric is utilized to quantify the overall resemblance in shape and temporal synchronization between trajectories. The HD serves to identify the maximal point of divergence. Concurrently, the RMSE provides a measure of the average pointwise discrepancy. 
The combined application of these metrics facilitates a comprehensive evaluation of trajectory prediction accuracy and similitude.

\subsection{Quantitative Analysis}

We evaluate the performance of \model{}-3B on both affordance perception and trajectory prediction tasks under in-domain and OOD settings.

In the in-domain experiment (Table~\ref{table1}), our model achieves an IoU score of 31.0 on the affordance perception task, substantially outperforming all baselines, including the much larger RoboBrain-7B (11.79), despite using only 50\% of the training data. 
This result highlights the effectiveness of our reward design in producing spatially grounded, semantically meaningful predictions under limited supervision, validating our motivation to align learning signals with task utility.
On the trajectory prediction task, \model{} also has strong performance, with an average distance error of 110.87, competitive with models trained with full supervision and larger capacities, demonstrating that our trajectory reward captures core requirements for physical and goal-directed plausibility, as intended.
In the OOD experiment (Table~\ref{table2}), \model{} continues to show robust generalization. 
It achieves the highest Grasp-IoU (34.65) and competitive performance on other UMD affordance subtasks. 
More notably, it achieves the lowest average trajectory error (131.99) on the VAIT benchmark, outperforming all open-source and supervised fine-tuned models, including Qwen2.5-VL-32B-Instruct and RoboBrain-7B.
These results collectively demonstrate \model{}’s ability to generalize under distribution shifts while maintaining high task accuracy across both perception and prediction.

\subsection{Qualitative Analysis}
\begin{figure}
    \centering
    \includegraphics[width=\linewidth]{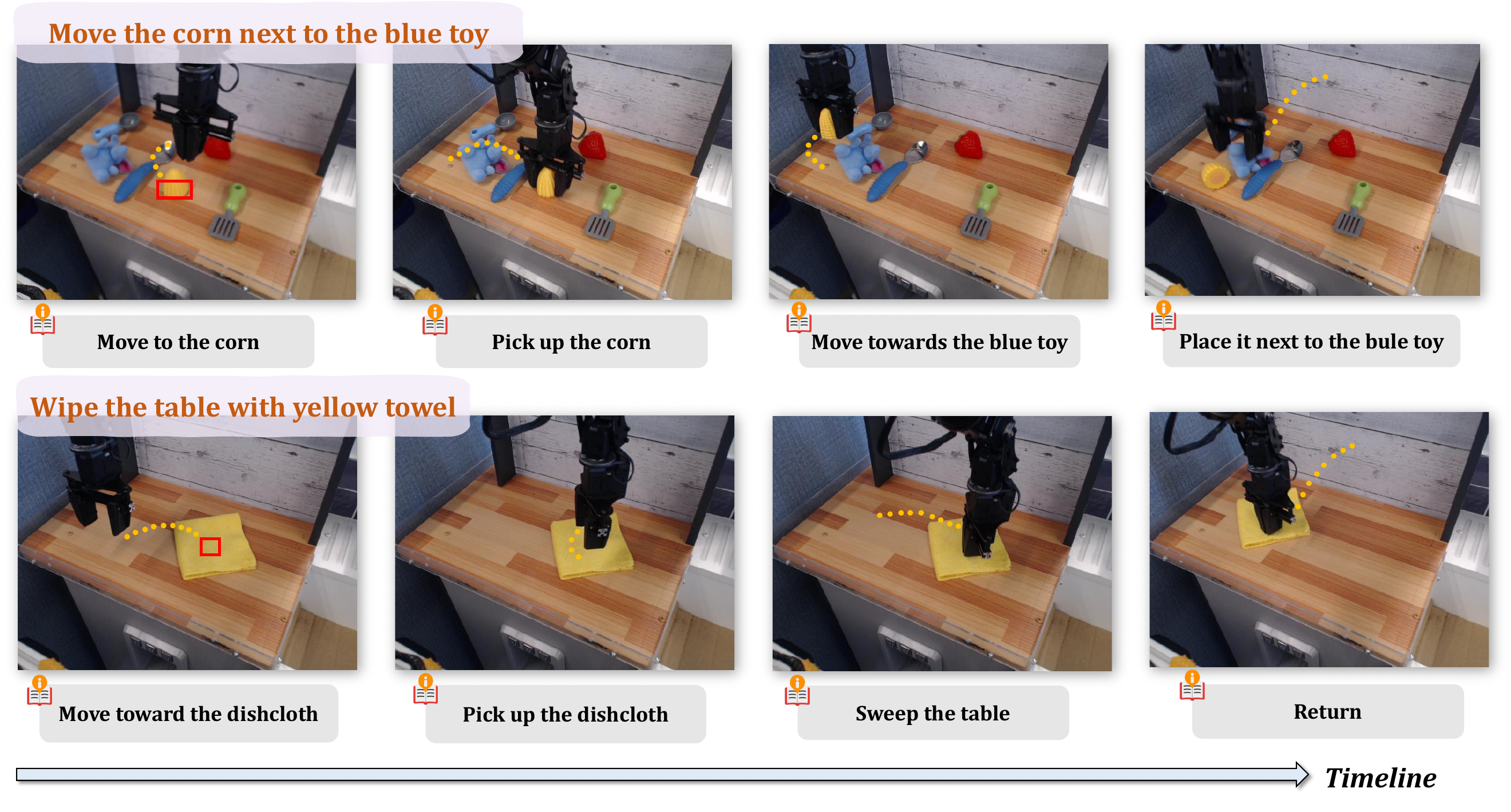}
     \caption{Illustration of integrated affordance perception and trajectory prediction. This figure presents examples where potential interaction regions (affordances) are identified, indicated by \textcolor{red}{red bounding boxes}, and a corresponding predicted action trajectory is shown as \textcolor{orange}{orange points}.}
    \label{fig:demo}
\end{figure}

\begin{figure*}[t]
\centering
\includegraphics[width=0.95\linewidth]{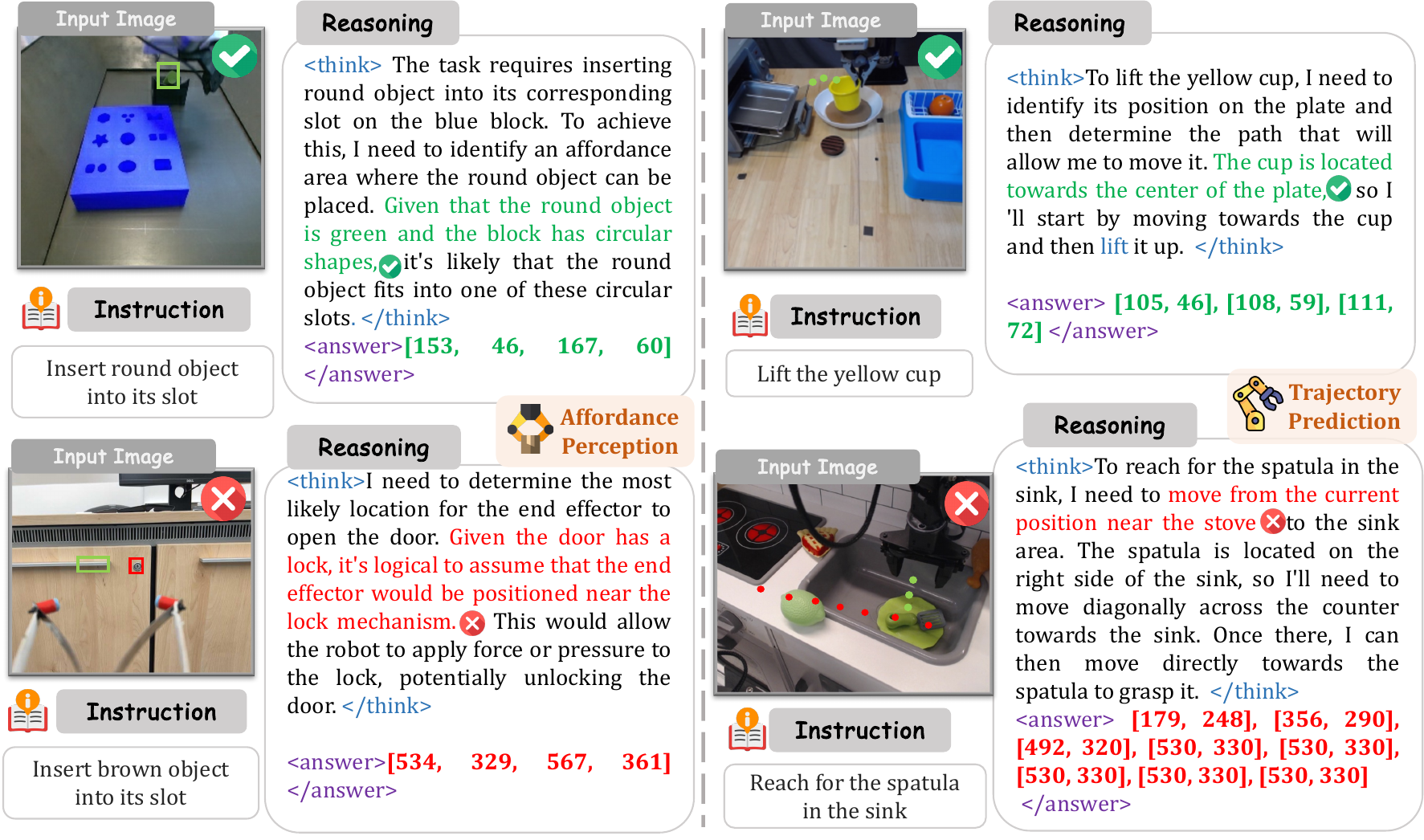}
\caption{Case studies illustrate both successful ``aha moments'' and failure cases in \model's predictions.
In each case, the final answer predicted by \model{} is enclosed in \texttt{<answer>} tags, while visualizations of the model’s prediction or ground truth are shown using \textcolor{green}{green boxes} and \textcolor{green}{green points}.
The model’s reasoning process is shown within \texttt{<think>} tags, where \textcolor{green}{green text} highlights key reasoning steps in successful cases, and \textcolor{red}{red text} marks the part responsible for the final failure. }
\label{fig:casestudy} 
\vspace{-3mm}
\end{figure*}

To qualitatively assess the capabilities of \model, we present visual examples of its performance in affordance perception and trajectory prediction. 
To better elucidate the practical significance of \model{} in real-world physical interactions, Figure \ref{fig:demo} demonstrates its engagement of spatial perception and trajectory prediction capabilities in response to two high-level human instructions: ``Move the corn next to the blue toy'' and ``Wipe the table with yellow towel''.
Furthermore, we extend the qualitative analysis to more nuanced reasoning processes.
As shown in Figure~\ref{fig:casestudy}, we observe an ``aha moment'' where \model{} demonstrates emergent reasoning abilities without any CoT data.
Take the first case, for example: when instructed to insert a round object into its slot, the model first identifies the object’s attributes (color and shape), then locates the matching slot—indicating implicit planning.
This suggests that \model{} can spontaneously perform multi-step reasoning in structured tasks, consistent with recent findings on emergent behavior in LLMs~\citep{DeepSeek-R1}.

We also present error analysis in Figure~\ref{fig:casestudy}. 
First, we identify instances of reasoning hallucinations during the model's inference process, primarily stemming from a lack of commonsense knowledge. These hallucinations can accumulate over steps and ultimately lead to incorrect final predictions.
In the first failure case, the model lacks commonsense understanding of uncommon doorknobs in the physical world. As a result, it fails to recognize the doorknob as an interactable region relevant to the instruction and instead incorrectly focuses on the lock in the image. This leads to a faulty reasoning path: ``the end effector approaching the keyhole will open the door.''
In another failure case, we observe that the model sometimes misperceives the gripper’s spatial position, which negatively impacts its reasoning.
For example, the model mistakenly believes the gripper is next to the stove. Based on this incorrect perception, it generates a trajectory that moves from the stove to the right side of the sink—resulting in an action that deviates from the intended goal.
These failure cases suggest that while \model excels at structured tasks, it still lacks mechanisms for verifying its internal beliefs or correcting perceptual errors.
Future work could explore incorporating external commonsense knowledge bases or self-monitoring modules to improve robustness in ambiguous or unfamiliar scenarios.



\subsection{Analysis of Reward Design}

\begin{wrapfigure}{r}{0.4\textwidth} 
    \centering
    \includegraphics[width=0.4\textwidth]{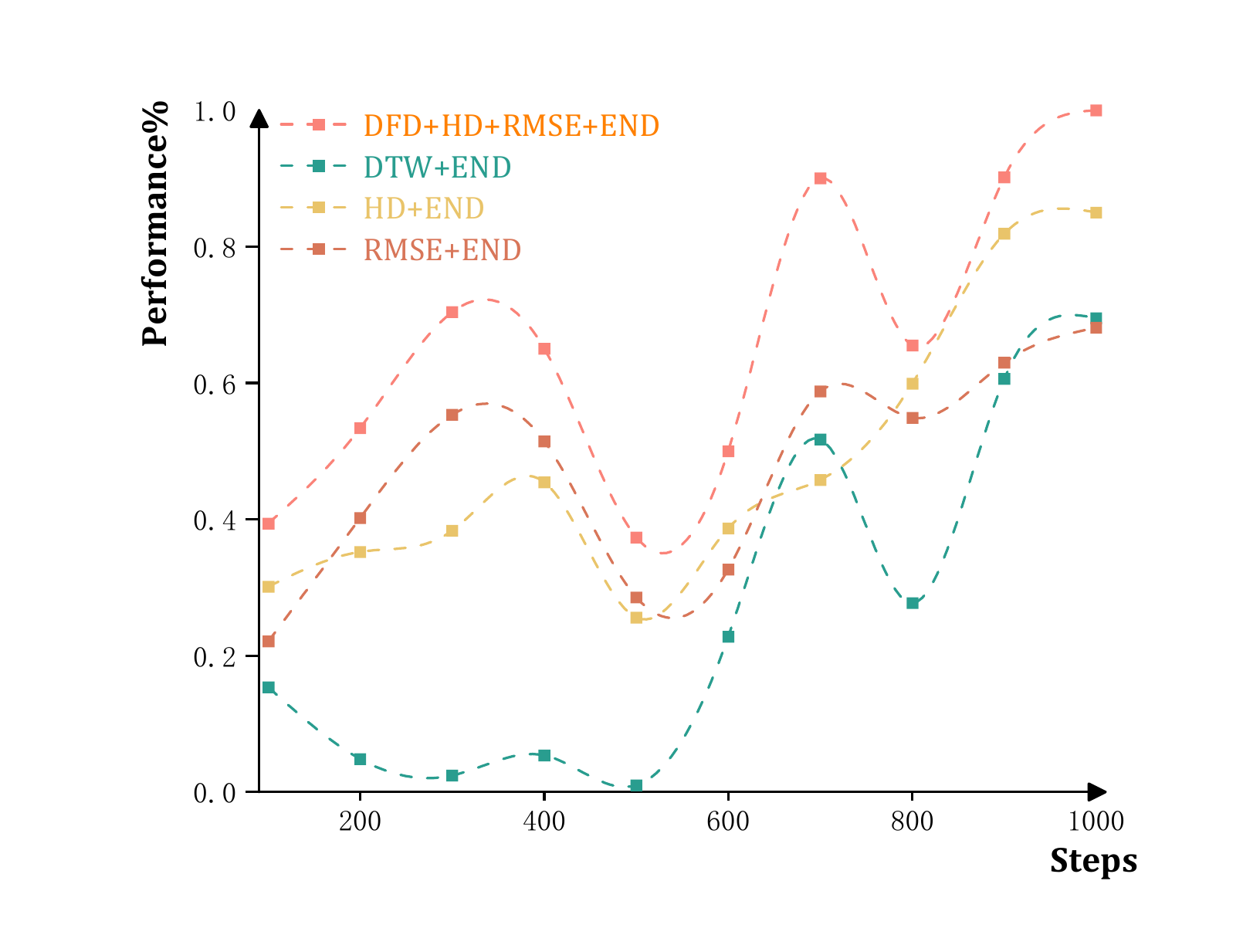} 
    \caption{Performance comparison across different reward designs.}
    \label{fig:reward_analysis}
    \vspace{-3mm}
\end{wrapfigure}

To better understand the impact of different reward components on model performance, we conduct an ablation study using the Trajectory Match Reward in Equation~\ref{rrr} as an example. 
As shown in Figure~\ref{fig:reward_analysis}, we compare several variants of the reward formulation by evaluating their influence on task performance over training steps.
Since distance-based metrics (e.g., DFD, HD, RMSE, Endpoint Error) are better when lower, we normalize and negate these values to convert them into a unified performance metric, which is used as the vertical axis in the plot.

The orange curve represents the full reward configuration that combines DFD, HD, RMSE, and Endpoint Error, which consistently achieves the highest performance across most training stages. 
This indicates that aggregating multiple complementary metrics provides a more reliable and stable learning signal for trajectory alignment.
The light blue curve (DTW + Endpoint) corresponds to a method that combines dynamic time warping (DTW)~\citep{muller2007dynamic} with endpoint matching. 
This method performs the worst, indicating that DTW is less effective for spatial reasoning in physical tasks.
The green (HD + Endpoint) and purple (RMSE + Endpoint) curves show moderate performance, suggesting that single metrics capture limited aspects of trajectory quality compared to the full reward.
These results highlight the importance of multi-metric reward design in reinforcement learning for robotic manipulation. By integrating diverse geometric and temporal criteria, the full reward formulation enables the agent to develop a deeper understanding of spatial structure and action consistency.


\section{Conclusion}

In this paper, we present \model, a novel reinforcement learning framework that advances robotic manipulation by replacing expensive human supervision with verifiable, rule-based rewards.
By introducing task-aligned reward functions, including Affordance Perception and Trajectory Match, our method enables more systematic physical reasoning and improved generalization, particularly in out-of-domain scenarios.
Unlike prior approaches that depend heavily on annotated data, \model  ~leverages immediate, structured feedback to guide learning, leading to shorter inference paths and improved computational efficiency.
Our experiments demonstrate that \model ~not only reduces supervision costs but also achieves substantial performance gains with limited training data, establishing it as a promising paradigm for scalable and adaptable robot learning.
We provide a detailed discussion of limitations in Appendix~\ref{limitation}.

\bibliographystyle{rusnat}
\bibliography{custom}

\appendix

\section{Limitation}
\label{limitation}
While \model{} demonstrates significant advancements in robotic manipulation through reinforcement learning with verifiable rewards, two limitations should be acknowledged.
The current framework focuses on affordance perception and 2D trajectory prediction. Real-world robotic manipulation often involves more intricate, long-horizon tasks that require a broader range of skills, such as fine-grained force control, manipulation of deformable objects, and dynamic interaction with the environment. We anticipate that future access to larger-scale, more comprehensive expert datasets will facilitate the application of our proposed paradigm to address these complex real-world challenges.
Meanwhile, the translation of predicted 2D image-space trajectories to precise and safe 3D robot movements in physical environments also presents challenges not fully captured by the existing evaluation metrics. However, recent advancements in 3D Shape Understanding and Reconstruction from 2D Images\citep{wen20223d,yin2021learning,xue2023ulip,liao2022kitti}, which involve training deep learning models to directly predict the six-degree-of-freedom (6DoF) pose (i.e., position and orientation) and three-dimensional shape of objects from either 2D images or point cloud data, have demonstrated considerable potential for directly translating 2D projections into real-world applications.

\end{document}